\documentclass[twocolumn, switch]{article} % Method A for two-column formatting
\usepackage{preprint}
\usepackage{amsmath,amssymb,amsfonts}
\usepackage{algpseudocode}
\usepackage{graphicx}
\usepackage{comment}
\usepackage{url} 
\usepackage{textcomp}
\usepackage{listings}
\usepackage{algorithm}
\usepackage{url}            % simple URL typesetting
\usepackage{booktabs}       % professional-quality tables
\usepackage{nicefrac}       % compact symbols for 1/2, etc.
\usepackage{microtype}      % microtypography
\usepackage{lipsum}
\setupPageNumbers
\usepackage{rotating}

\usepackage{lscape}
\usepackage{multirow}
\usepackage{float}
\usepackage{tabularray}
\usepackage{adjustbox}
\usepackage{tabularx,ragged2e,booktabs}
\usepackage{tabulary,booktabs}
\usepackage{tabularray}
\usepackage{academicons}
\usepackage{graphicx} % For including images
\usepackage{lipsum}   % For dummy text, if needed

\usepackage{amsmath, amsthm, amssymb, amsfonts}

\usepackage[numbers,sort&compress]{natbib}
\usepackage[utf8]{inputenc}	% allow utf-8 input
\usepackage[T1]{fontenc}	% use 8-bit T1 fonts
\usepackage{xcolor}		% colors for hyperlinks
\usepackage[colorlinks = true,
            linkcolor = purple,
            urlcolor  = blue,
            citecolor = cyan,
            anchorcolor = black]{hyperref}	% Color links to references, figures, etc.
\usepackage{booktabs} 		% professional-quality tables
\usepackage{nicefrac}		% compact symbols for 1/2, etc.
\usepackage{microtype}		% microtypography
\usepackage{lineno}		% Line numbers
\usepackage{float}			% Allows for figures within multicol
\usepackage{lipsum}		%  Filler text

\usepackage{newfloat}
\DeclareFloatingEnvironment[name={Supplementary Figure}]{suppfigure}
\usepackage{sidecap}
\sidecaptionvpos{figure}{c}

\usepackage{titlesec}
\titlespacing\section{0pt}{12pt plus 3pt minus 3pt}{1pt plus 1pt minus 1pt}
\titlespacing\subsection{0pt}{10pt plus 3pt minus 3pt}{1pt plus 1pt minus 1pt}
\titlespacing\subsubsection{0pt}{8pt plus 3pt minus 3pt}{1pt plus 1pt minus 1pt}

\usepackage{tikz,xcolor,hyperref}

\definecolor{lime}{HTML}{A6CE39}
\DeclareRobustCommand{\orcidicon}{
	\begin{tikzpicture}
	\draw[lime, fill=lime] (0,0)
	circle [radius=0.16]
	node[white] {{\fontfamily{qag}\selectfont \tiny ID}};
	\draw[white, fill=white] (-0.0625,0.095)
	circle [radius=0.007];
	\end{tikzpicture}
	\hspace{-2mm}
}
\foreach \x in {A, ..., Z}{\expandafter\xdef\csname orcid\x\endcsname{\noexpand\href{https://orcid.org/\csname orcidauthor\x\endcsname}
			{\noexpand\orcidicon}}
}
\newcommand{\custombio}[3]{
    \begin{minipage}[t]{\columnwidth}
    \vspace{0pt} % Ensures top alignment with the text
    \parbox[t]{1in}{%
        \vspace{0pt} % Ensures top alignment
        \includegraphics[width=1in,height=1.25in,clip,keepaspectratio]{#1}%
    }
    \hspace{5mm} % Space between image and text
    \begin{minipage}[t]{0.65\columnwidth}
    \vspace{0pt} % Ensures top alignment
    {\textsf{\textcolor{blue}{#2}}} % Author name in blue and sans-serif font
    {\textsf{\textcolor{black}{#3}}} % Biography text in black and same sans-serif font
    \end{minipage}
    \end{minipage}
    \vspace{5mm} % Adds some space below the biography
}

\title{MM-SurvNet: Deep Learning-Based Survival Risk Stratification in Breast Cancer Through Multimodal Data Fusion}

\usepackage{authblk}

\author[1]{Raktim Kumar Mondol}
\author[2]{Ewan K.A. Millar}
\author[1]{Arcot Sowmya}
\author[1,*]{Erik Meijering} 
\affil[1]{School of Computer Science and Engineering, University of New South Wales, Sydney, Australia}
\affil[2]{Department of Anatomical Pathology, NSW Health Pathology, St. George Hospital}
\affil[*]{Correspondence: \texttt{erik.meijering@unsw.edu.au}}

\begin{document}

\twocolumn[ % Method A for two-column formatting
\begin{@twocolumnfalse} % Method A for two-column formatting

\maketitle
\begin{abstract}
Survival risk stratification is an important step in clinical decision making for breast cancer management. We propose a novel deep learning approach for this purpose by integrating histopathological imaging, genetic and clinical data. It employs vision transformers, specifically the MaxViT model, for image feature extraction, and self-attention to capture intricate image relationships at the patient level. A dual cross-attention mechanism fuses these features with genetic data, while clinical data is incorporated at the final layer to enhance predictive accuracy. Experiments on the public TCGA-BRCA dataset show that our model, trained using the negative log likelihood loss function, can achieve superior performance with a mean C-index of 0.64, surpassing existing methods. This advancement facilitates tailored treatment strategies, potentially leading to improved patient outcomes.
\end{abstract}
 % (optional)
\vspace{0.35cm}
\end{@twocolumnfalse} % Method A for two-column formatting
] % Method A for two-column formatting

%\begin{multicols}{2} % Method B for two-column formatting (doesn't play well with line numbers), comment out if using method A

%%%%%%%%%%%%%%%  Main text   %%%%%%%%%%%%%%%
% \linenumbers
%\begin{comment}

\section{Introduction}
\begin{figure}[t]
\centering
\includegraphics[width=0.49\textwidth]{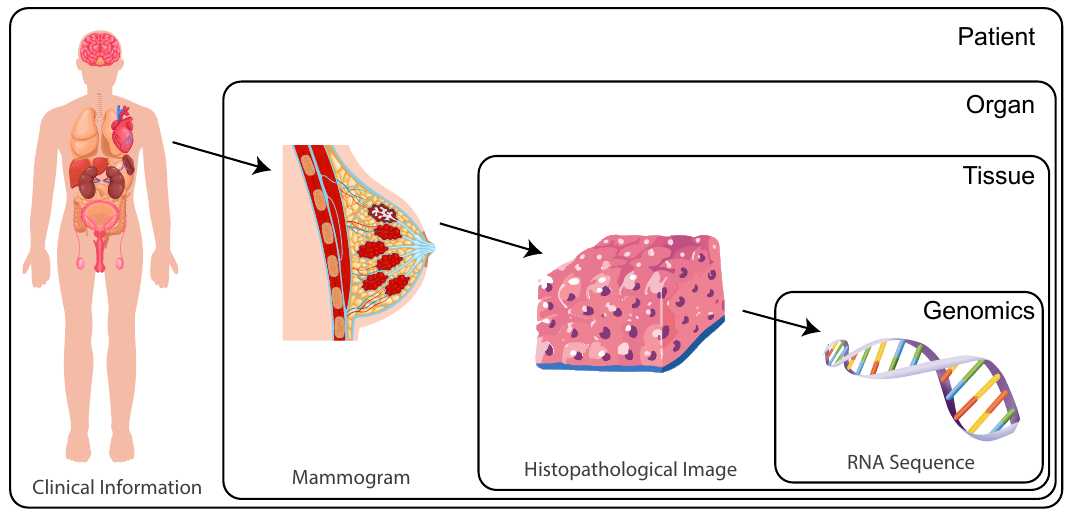}

\caption{Multimodal Data for Breast Cancer Characterization: This figure illustrates the  diverse data types at multiple biological levels to comprehensively characterize breast cancer. It encompasses clinical data at the patient level, mammogram images at the organ level, histopathology images at the tissue level, and gene expression data (e.g., RNA Sequencing) at the molecular level, thereby providing a holistic view of the cancer's nature and behavior. }
\label{fig:multimodaldata}
\end{figure}

{T}{he} inherent heterogeneity of breast cancer poses challenges for prediction of prognosis and treatment decisions particularly in post-menopausal estrogen receptor positive (ER+) breast cancer, with previous studies reporting conflicting results on the survival difference between Luminal A and B metastatic breast cancer patients~\cite{10.1186/s12885-022-10098-1, 10.1016/j.breast.2020.09.006}. A common critical clinical dilemma is the selection of those early ER+ breast cancer patients at high risk of recurrence who may benefit from the addition of chemotherapy to endocrine therapy. Traditional risk prediction methods, based mainly on clinicopathological factors, may not entirely account for the complex biology of cancer~\cite{10.7717/peerj.12202}. To address this issue, molecular markers and gene expression profiles have emerged as a potential addition  to current decision making tools~\cite{10.1371/journal.pcbi.1008133}. Therefore, combining histopathology image, molecular and clinicopathological factors (see Figure \ref{fig:multimodaldata} adapted from \cite{Subramanian2021}) into a single risk prediction model could potentially enhance risk prognostication in the clinic~\cite{10.3389/fonc.2021.612450}. Here we propose a multimodal risk prediction model for ER+ breast cancer and compare its performance to existing methods. The significance of our work lies in potentially refining ER+ breast cancer survival risk prediction, facilitating informed therapeutic decision making~\cite{10.3389/fgene.2021.709027}. 

%\begin{comment}

\begin{figure*}[t]
\centering
\includegraphics[width=0.95\textwidth]{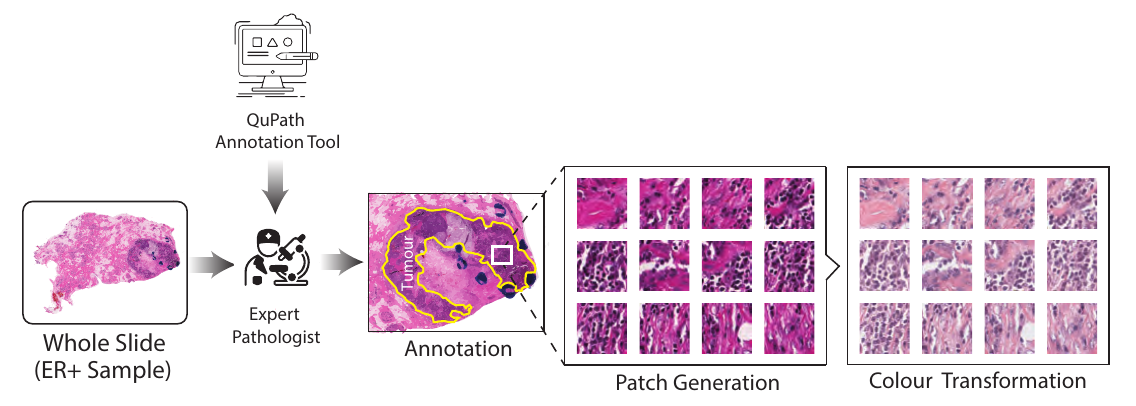}
\caption{Preprocessing Methodology for Histopathology Images: Whole Slide Images (WSIs) are first annotated using the QuPath annotation tool by expert pathologists. From these annotated regions, non-overlapping patches are systematically extracted. Subsequently, a color transformation is applied to ensure uniformity in color representation across datasets.}
\label{fig:preprocessing}
\end{figure*}
\section{Related Works}
\label{sec:format}
Breast cancer accounts for about 30\% of all cancer cases in women, with varying incidence rates across regions and notably higher rates in developed countries~\cite{10.1136/bmjopen-2022-061205}. The ER status, specifically ER+ breast cancer, which encompasses Luminal A and Luminal B subtypes, is crucial in shaping treatment strategies and predicting prognosis~\cite{10.7717/peerj.12202}. While Luminal A tumours are usually low-grade with a favorable prognosis (Ki-67 \textless{} 14\%), Luminal B tumours are usually higher grade and pose a higher recurrence risk and worse outcome (Ki-67 $\geq$ 14\%)~\cite{10.7717/peerj.12202, 10.1186/s12880-017-0239-z}. Traditional prediction models primarily centred on clinical and pathological factors fall short, especially in capturing the intricacies and heterogeneity of ER+ breast cancer subtypes~\cite{10.1371/journal.pcbi.1009495}. Transitioning from traditional methods, early research in survival risk prediction using histopathology images shifted from reliance on hand-crafted features like texture and shape to leveraging deep learning for automatic extraction of complex, high-level features, resulting in enhanced accuracy~\cite{10.1038/s41598-022-19112-9}. In recent developments, multimodal approaches that combine imaging, clinical and molecular data have been shown to enhance prediction accuracy~\cite{10.1111/cas.15592, 10.3389/fcell.2021.720110, 10.1093/bioinformatics/btad025, 10.3389/fonc.2022.745258, 10.1109/isbi48211.2021.9434033, 10.1038/s43018-022-00416-8}. Comprehensive multimodal approaches, especially those that fuse whole slide images, clinical data and genetic information, could pave the way for superior predictive models.

\section{Data Preparation}
\label{sec:pagestyle}
Our study uses the hematoxylin and eosin (H\&E)-stained formalin-fixed paraffin-embedded (FFPE) digital slides of The Cancer Genome Atlas Breast Cancer (TCGA-BRCA) dataset \cite{TCGA-BRCA}, sourcing 249 whole-slide images (WSIs) from the Genomic Data Commons (GDC) Portal with a focus on Luminal A (149 samples) and Luminal B (100 samples) molecular subtypes. Slide annotations were obtained manually by an expert breast pathologist using QuPath \cite{Bankhead2017}, prioritising tumour localisation including stroma and tumour-infiltrating lymphocytes (TILs) and excluding necrotic areas. H\&E-stained tissues underwent downsampling to a resolution of 0.25 µm/pixel, with tissue masks created to exclude artifacts and non-tissue sections, and tumour regions split into $224\times224$ pixel patches. To counteract staining inconsistencies, singular value decomposition-based normalisation was applied on the images (See Figure~\ref{fig:preprocessing}) using  Macenko method \cite{Macenko2009}. Furthermore, we processed RNA-sequencing data with RSEM and selected PAM50 genes for analysis, given their significance in breast cancer subtyping~\cite{Li2011, 10.1200/jco.2008.18.1370}. Pertinent clinical variables, encompassing tumour grade, size, patient age and lymph node status, were meticulously prepared to enhance the predictive accuracy of the subsequent deep learning model evaluations. For experimental purposes, we adopted a five-fold cross-validation strategy, allocating 200 samples for training and 50 for validation while maintaining the distribution of Luminal subtypes and survival statuses.
\begin{figure*}[t]
\centering
\includegraphics[width=0.9\textwidth]{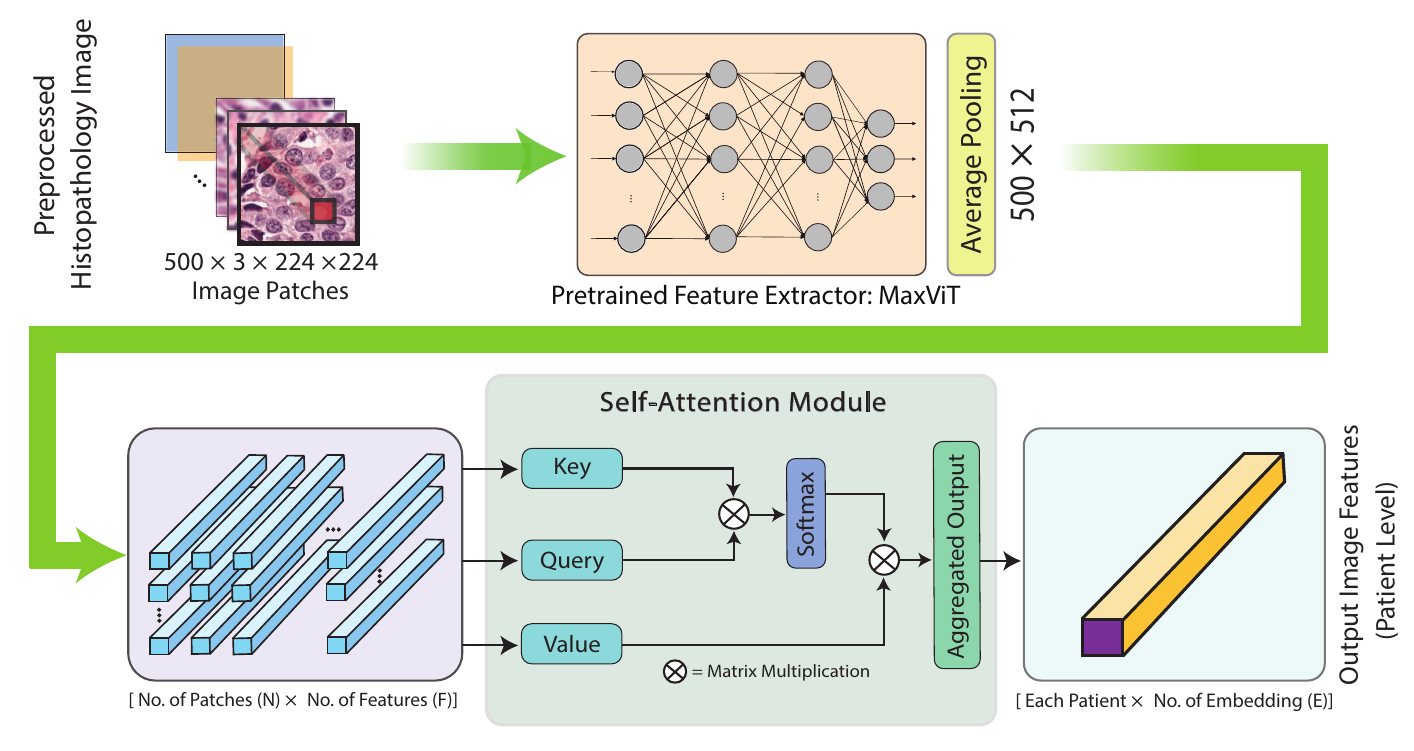}
\caption{The proposed MM-SurvNet architecture. It utilises a pretrained MaxViT to extract features from histopathology images, which are then processed through a self-attention network to aggregate patch-level features into a patient-level representation.}
\label{fig:architecture1}
\end{figure*}

\section{Proposed Method}
\label{sec:majhead}
\subsection{Histopathological Feature Extraction}

We utilised vision transformers (ViTs) and convolutional neural networks (CNNs) to extract features from histopathological images. Specifically, the MaxViT model, a prominent vision transformer, was employed for its effectiveness in capturing global image information and leveraging self-attention for comprehensive tumour analysis~\cite{tu2022maxvit}. For the purpose of comparison, ResNet50, a CNN-based model, was selected to extract local patterns within the images~\cite{10.1111/his.13844}.

\subsection{Patch-to-Patient Aggregation Using Self-Attention}

In our framework, the self-attention mechanism crucially aggregates patch-level features into a holistic patient-level representation. Notably, the same image patches act as the Key (K), Query (Q), and Value (V) within the attention paradigm, aiming to discern intricate interdependencies among patches from a singular histopathological image.

Mathematically, self-attention processes a sequence of image patches \( \mathbf{K}_1, \mathbf{Q}_1, \mathbf{V}_1, \ldots, \mathbf{K}_n, \mathbf{Q}_n, \mathbf{V}_n \) to yield a patient-level feature vector \( \mathbf{y} \). The mechanism computes a weighted sum of the Value vectors \( \mathbf{V}_j \) based on attention scores derived from the scoring function \( s(\mathbf{Q}_i, \mathbf{K}_j) \):

\begin{equation}
\mathbf{y}_i = \sum_{j=1}^{n} \text{softmax}\left(s(\mathbf{Q}_i, \mathbf{K}_j)\right) \mathbf{V}_j.
\end{equation}
In this formulation, the softmax function normalises scores, facilitating the model's emphasis on specific patches. The resultant \( \mathbf{y}_i \) is a weighted blend of patch-level Value vectors \( \mathbf{V}_j \), representing the patient-level feature.

Two primary advantages emerge:
\begin{enumerate}
    \item \textbf{Inherent Feature Importance:} Using image patches for Key, Query and Value enables the model to naturally highlight significant regions in histopathological images, capturing inherent sample heterogeneity.
    \item \textbf{Inter-Patch Contextualisation:} The mechanism grasps the context among distinct patches, contextualising each patch with respect to others for a comprehensive patient-level representation.
\end{enumerate}

The enriched patient-level vector, emphasising both context and focus, is pivotal for subsequent survival risk prediction tasks.

\begin{table*}[!t]
\centering
\caption{Performance Comparison using C-index on TCGA Data.}
\label{tab:performance_comparison}
\begin{tabular}{@{}ccccccccc@{}}
\toprule
CV Fold & \multicolumn{2}{c}{Multimodal (Imaging+Genetic+Clinical)} & \multicolumn{2}{c}{Imaging+Genetic} & \multicolumn{2}{c}{Imaging} & \multicolumn{2}{c}{Clinical} \\
\cmidrule(r){2-3} \cmidrule(r){4-5} \cmidrule(r){6-7} \cmidrule(r){8-9}
& MaxViT & ResNet50  & MaxViT & ResNet50 & MaxViT & ResNet50 & \multicolumn{2}{c}{CoxPH} \\
\midrule
1    & 0.70       & 0.64            & 0.72   & 0.61     & 0.45   & 0.47     & \multicolumn{2}{c}{0.50}  \\
2    & 0.57       & 0.49            & 0.44   & 0.58     & 0.47   & 0.47     & \multicolumn{2}{c}{0.24}  \\
3    & 0.66         & 0.62            & 0.63   & 0.65     & 0.55   & 0.57     & \multicolumn{2}{c}{0.34}  \\
4    & 0.62         & 0.55            & 0.50    & 0.48     & 0.60    & 0.60     & \multicolumn{2}{c}{0.57}  \\
5    & 0.70          & 0.50             & 0.57   & 0.51     & 0.62   & 0.65     & \multicolumn{2}{c}{0.71}  \\
\midrule
Mean    & $0.64\pm 0.06$     & $0.56\pm 0.06$          & $0.57\pm 0.10$  & $0.56\pm0.06$    & $0.53\pm0.08$  & $0.55\pm0.08$    & \multicolumn{2}{c}{$0.47\pm0.17$}  \\

\bottomrule
\end{tabular}
\end{table*}

\begin{figure*}[t]
\centering
\includegraphics[width=1\textwidth]{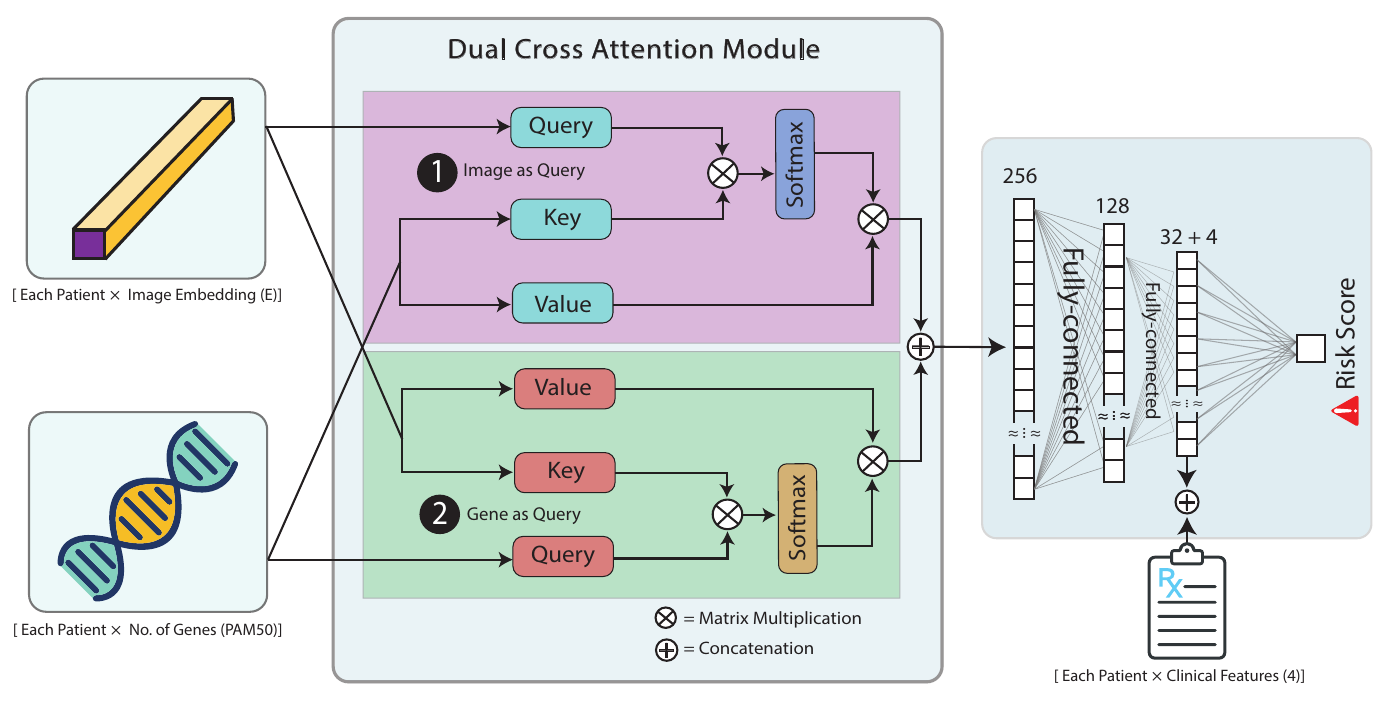}
\caption{The proposed MM-SurvNet architecture employs a multimodal deep learning framework for survival risk prediction in cancer patients. Image embedding is concatenated with genetic data using dual cross-attention mechanisms and further integrated with clinical data in the final layer.}
\label{fig:architecture2}
\end{figure*}

\subsection{Multimodal Fusion Using Dual Cross-Attention}
Fusing data from diverse sources is paramount in multimodal learning. In our architecture, we deploy a dual cross-attention mechanism to achieve a refined interaction between histopathology image features and genetic expression profiles. Specifically, this encompasses two attention operations:

\begin{enumerate}
    \item Image features serve as the Query (Q) while genetic features act as both the Key (K) and Value (V).
    \item Conversely, genetic features form the Query, with image features becoming the Key and Value.
\end{enumerate}

Let \( \mathbf{I} = \{\mathbf{i}_1, \mathbf{i}_2, \ldots, \mathbf{i}_n\} \) be the set of image feature vectors and \( \mathbf{G} = \{\mathbf{g}_1, \mathbf{g}_2, \ldots, \mathbf{g}_m\} \) denote the genetic feature vectors. The enriched feature vectors for the \( k^\text{th} \) image and genetic patches are:
\[
\mathbf{y}^I_k = \sum_{j=1}^{m} \text{softmax}\left( s(\mathbf{i}_k, \mathbf{g}_j) \right) \mathbf{g}_j,
\]
\[
\mathbf{y}^G_k = \sum_{i=1}^{n} \text{softmax}\left( s(\mathbf{g}_k, \mathbf{i}_i) \right) \mathbf{i}_i.
\]
Here, the scoring function \( s(.) \) gauges the relevance between Queries and Keys. These weighted sums, after being normalised by the softmax function, are then concatenated to form a unified feature representation:
\[
\mathbf{y}_k = \text{Concatenate}(\mathbf{y}^I_k, \mathbf{y}^G_k)
\]

The dual cross-attention offers two primary advantages:

\begin{enumerate}
    \item \textbf{Bidirectional Contextualisation:} The mechanism captures intermodal dependencies in both directions---image to genetic and vice versa---providing a holistic contextualised representation.
    
    \item \textbf{Enhanced Selective Attention:} By alternating the role of Query, the architecture gleans specific features from both modalities that are crucial for each other, ensuring compact and informative fused feature vectors.
\end{enumerate}

The resultant vectors \( \mathbf{y}_k \) are thus contextually enriched and optimised for tasks such as survival risk stratification.

\subsection{MM-SurvNet Model Training}
The MM-SurvNet model (Fig.~\ref{fig:architecture1}) employs either a pretrained Vision Transformer (MaxViT) or a CNN (ResNet50) for histopathological image feature extraction. The multimodal data is subsequently incorporated into the MM-SurvNet model (Fig.~\ref{fig:architecture2}) to enhance prediction accuracy. The Adam optimiser with decoupled weight decay (AdamW) with a learning rate of 0.001 and a mini-batch size of 12 governs the optimisation, which includes an early stopping criterion set at a patience of 5 and a maximum of 150 epochs~\cite{loshchilov2018decoupled}. The primary loss function is the negative log-likelihood (NLL), aligning with our survival analysis objectives, and optimisation targets the concordance index (C-index) as the key performance metric\cite{LONGATO2020103496}. The NLL loss is calculated as:
\begin{equation}
\label{loss}
\text{Loss} = - \frac{\sum_{i} \left( \log(h_i) - \log \left( \sum_{j \in R_i} \exp(\log(h_j)) \right) \right) \cdot d_i}{\sum_{i} d_i}
\end{equation}
where \( h = \exp(\log(h)) \) are the hazards, \( R \) is the risk set, and \( d \) represents the event. The optimisation, subject to early stopping criteria, ranged from 20 to 50 minutes on an NVIDIA Tesla V100 32GB GPU. The best model weights, based on the minimal NLL loss (\ref{loss}) were retained for testing.

\begin{figure}[!t]
\centering
\includegraphics[width=0.48\textwidth]{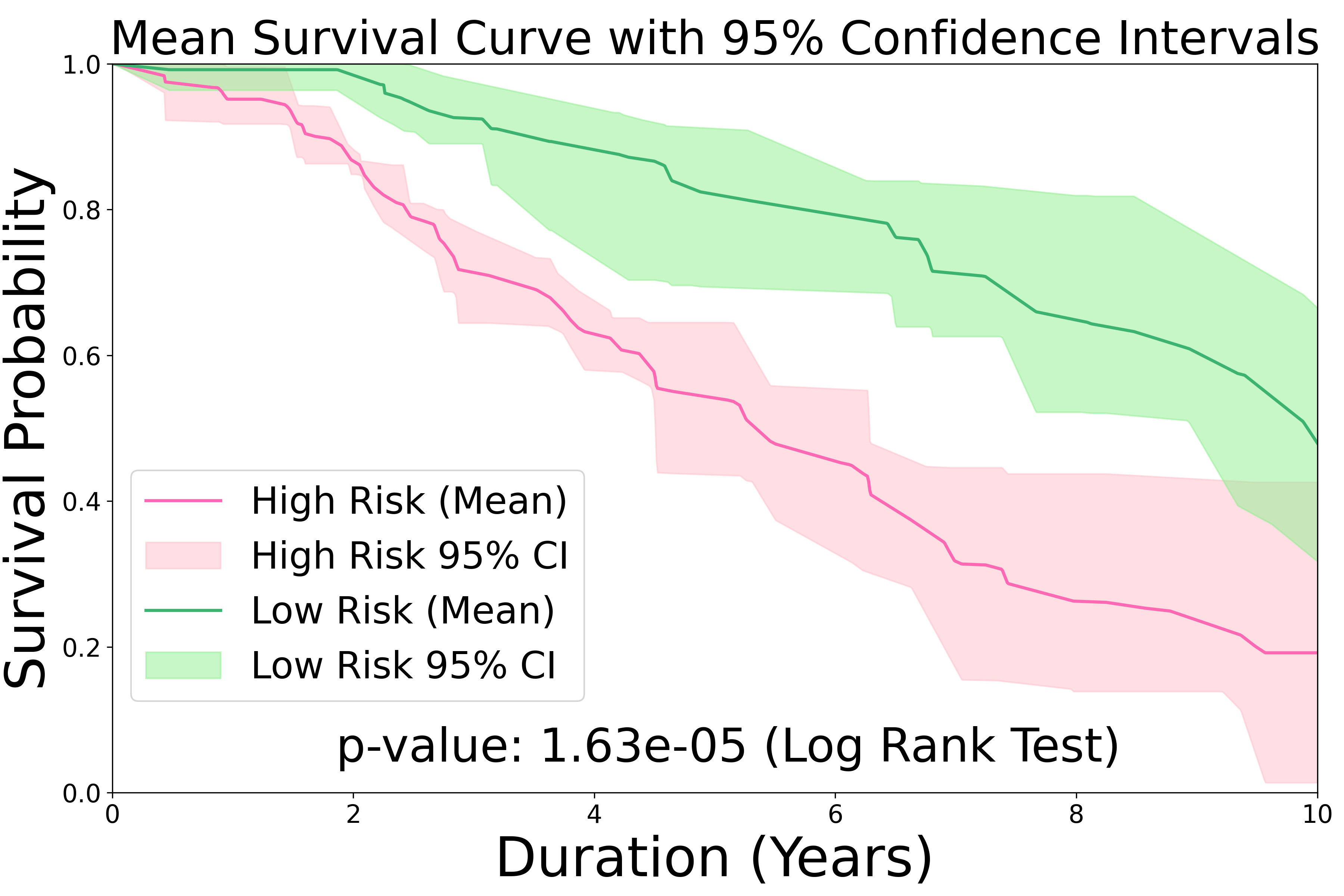}
\caption{Mean survival curves aggregated from all five cross-validation (CV) tests with 95\% confidence intervals. In the depicted curves, the log-rank test p$<$0.05.}
\label{fig:sc}
\end{figure}

\section{Experimental Results}
\label{sssec:subsubhead}

In our study on survival risk stratification for ER+ breast cancer patients, we integrated multimodal data---histopatholo\-gical images, genetic information, and clinical data---resulting in enhanced predictive accuracy. Our method achieved a notable C-index of 0.64, superior to models leveraging a single data modality (Table \ref{tab:performance_comparison}). For instance, models based on histopathological image data only or on clinical information only, yielded C-indices of 0.53 and 0.47, respectively. An ablation study (Table~\ref{tab:performance_comparison}) highlighted the significance of our holistic, multimodal approach. When integrating all feature sets, the MaxViT model surpassed ResNet50, emphasising the advantages of combining clinical, imaging and genetic data for more nuanced and accurate risk predictions. Kaplan-Meier survival analysis (Fig.~\ref{fig:sc}) further demonstrates our method’s efficacy in differentiating outcomes between risk groups (log-rank test p$<$0.05), with an Integrated Brier Score of 0.11 for 5 years and 0.16 for 10 years~\cite{JAGER2008560}.

\section{Discussion}
\label{sec:print}

The findings of our study assessing multimodal data integration for survival risk stratification in ER+ breast cancer patients underscore the superiority of multimodal approaches. Our proposed model, amalgamating histopathological images, genetic data and clinical parameters, achieved a significant C-index of 0.64, outshining single-modal counterparts. Specifically, the MaxViT architecture consistently surpassed ResNet50 in handling this multimodal data. While the Kaplan-Meier analysis reinforced our model's aptitude in patient risk stratification, there was notable performance variability across cross-validation folds, suggesting potential dataset nuances or imbalances. Moreover, the traditional CoxPH model's C-index of 0.47 highlighted the advantages of deep learning in capturing cancer prognosis' intricate nature. The analysis not only substantiated our model's proficiency in stratifying patients but also highlighted significant survival differences between the high-risk and low-risk groups, as revealed in the survival curve plots.

Future work will encompass broader data types and probe the model's utility across diverse cancer subtypes. While our study yielded promising results, it suffers from certain limitations. Our sample size, though adequate for the current modeling exercise, might benefit from expansion to ensure broader generalisability. The study's concentration on ER+ breast cancer means that its applicability to other cancer subtypes remains as yet untested. The variability in performance across different cross-validation folds suggests potential dataset imbalances or unidentified influencing factors. Lastly, the model's dependence on three types of data may pose challenges in settings where one or more of these data modalities is unavailable.

\section{Conclusion}
Our study highlights the significance of multimodal data integration in advancing survival risk stratification for ER+ breast cancer patients. This integrative methodology enhances prediction accuracy and sets the foundation for personalised treatment strategies in oncology.

\section*{Compliance with Ethical Standards}
This research study was conducted retrospectively using human subject data made available in open access by The Cancer Genome Atlas Breast Cancer (TCGA-BRCA) dataset, accessible through the National Cancer Institute's \href{https://portal.gdc.cancer.gov/projects/TCGA-BRCA}{Genomic Data Commons (GDC)} portal. Ethical approval was not required for this study, in accordance with the \href{https://www.cancer.gov/ccg/research/genome-sequencing/tcga/history/ethics-policies}{ethical policies} set forth by The Cancer Genome Atlas program.

\section*{Conflicts of Interest}
No funding was received for conducting this study. The authors have no relevant financial or non-financial interests to disclose.

\section*{Acknowledgement}
This research was undertaken with the assistance of resources and services from the National Computational Infrastructure (NCI), which is supported by the Australian Government. Additionally, data preprocessing was performed using the computational cluster Katana, which is supported by Research Technology Services at UNSW Sydney.
%%%%%%%%%%%%%%   Bibliography   %%%%%%%%%%%%%%
\footnotesize
\bibliography{references}

%%%%%%%%%%%%  Supplementary Figures  %%%%%%%%%%%%

\custombio{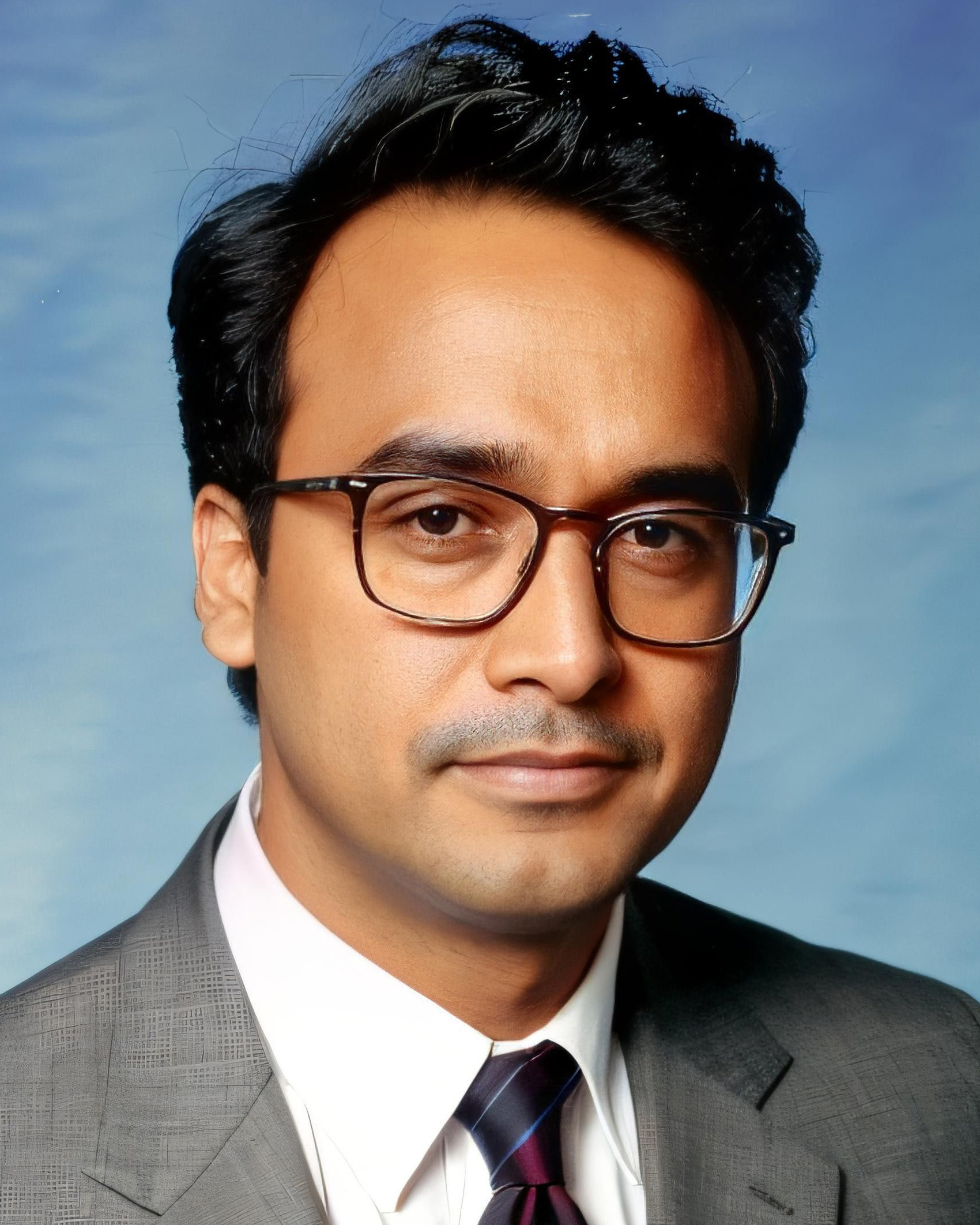}{\href{https://research.unsw.edu.au/people/mr-raktim-kumar-mondol}{Raktim Kumar Mondol}}{is a PhD candidate in Computer Science and Engineering, specializing in computer vision and bioinformatics. He completed his MEng in Engineering with High Distinction from RMIT University, Australia. Mondol's research interests include histopathological image analysis, clinical prognosis prediction, and enhancing clinical understanding through the interpretability of computational models.}

\custombio{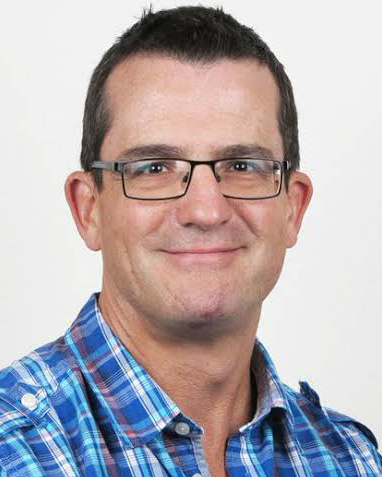}{\href{https://www.unsw.edu.au/staff/ewan-millar}{Ewan Millar}}{is a Senior Staff Specialist Histopathologist with NSW Health Pathology at St George Hospital Sydney with expertise in breast cancer pathology and translational research and a strong interest in AI and digital pathology applications.}

\custombio{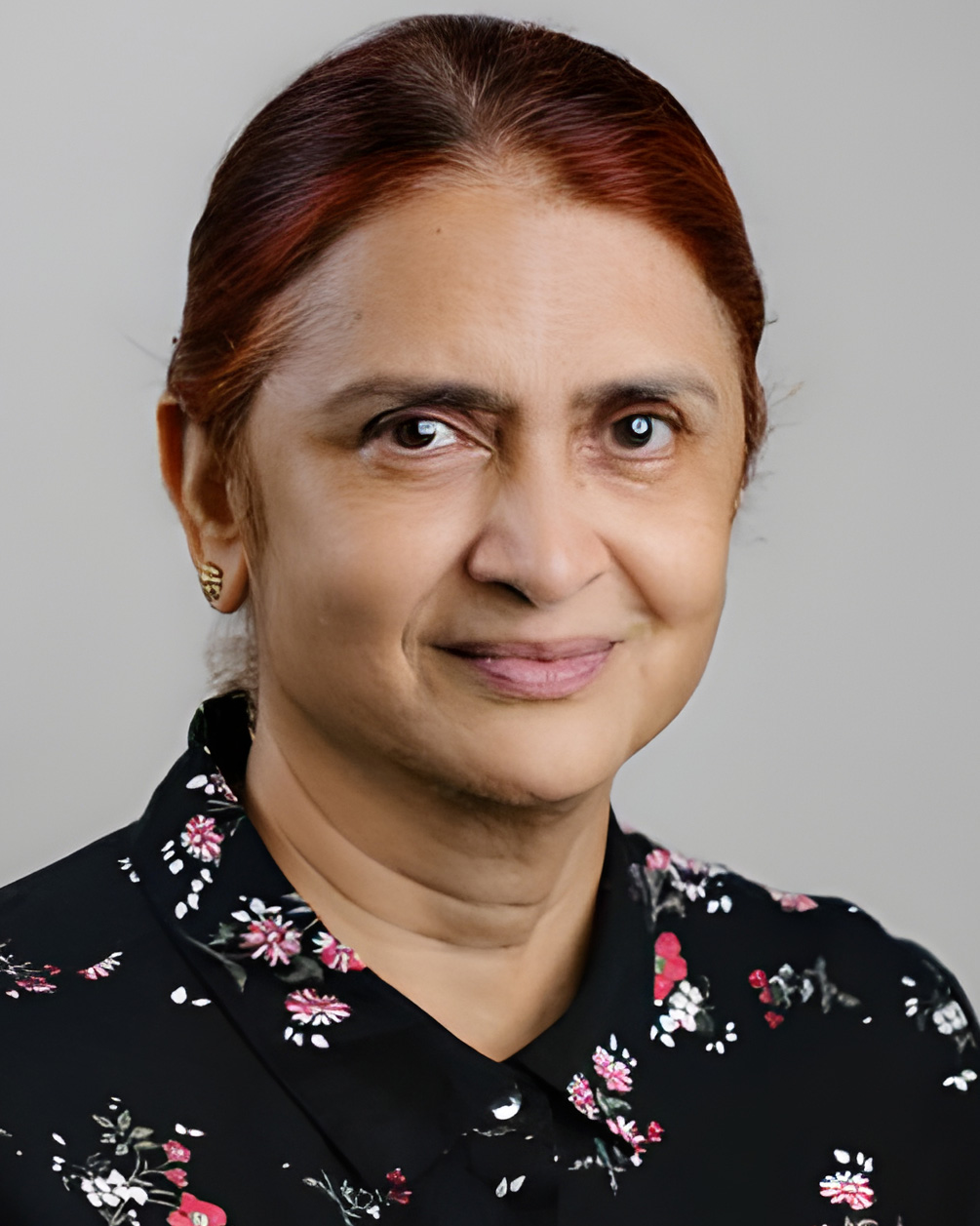}{\href{https://research.unsw.edu.au/people/professor-arcot-sowmya}{Arcot Sowmya}}{is Professor in the School of Computer Science and Engineering, UNSW. Her major research interest is in the area of Machine Learning for Computer Vision and includes learning object models, feature extraction, segmentation and recognition based on computer vision, machine learning and deep learning. In recent years, applications in the broader health area are a focus, including biomedical informatics and rapid diagnostics in the real world. All of these areas have been supported by competitive, industry and government funding.}

\custombio{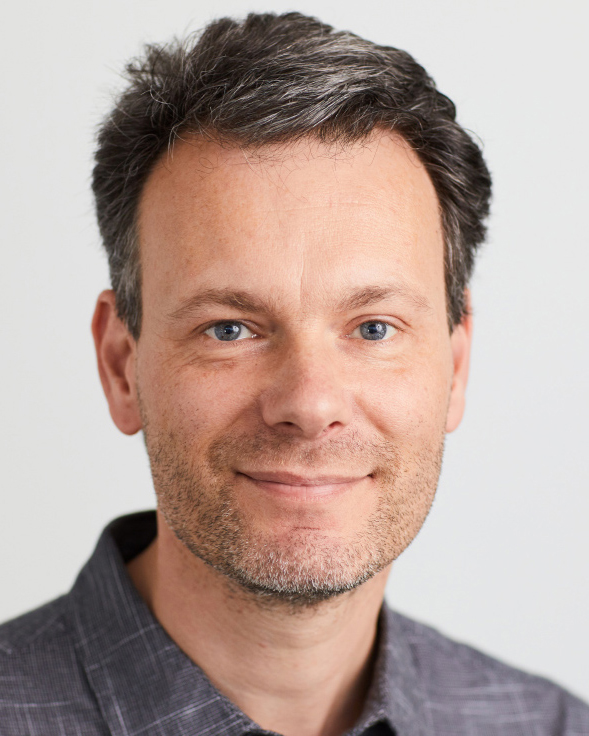}{\href{https://imagescience.org/meijering/}{Erik Meijering}}{(Fellow, IEEE), is a Professor of Biomedical Image Computing in the School of Computer Science and Engineering. His research focusses on the development of innovative computer vision and machine learning (in particular deep learning) methods for automated quantitative analysis of biomedical imaging data.}

%%%%%%%%%%%%%%%%   End   %%%%%%%%%%%%%%%%
%\end{multicols}  % Method B for two-column formatting (doesn't play well with line numbers), comment out if using method A
\end{document}